\documentclass[10pt,conference,compsocconf]{IEEEtran}

\usepackage{times}
\usepackage{epsfig}
\usepackage{graphicx}
\usepackage{amsmath}
\usepackage{amssymb}
\usepackage{booktabs}
\usepackage{rotating}
\usepackage{caption}
\usepackage{subcaption}
\usepackage{xcolor}
\usepackage{soul}
\usepackage{threeparttable}
\usepackage{multirow}
\usepackage[
  breaklinks=true,
  colorlinks,
  bookmarks=false]{hyperref}
\usepackage{cleveref}
\usepackage{cite}

\def\IEEEbibitemsep{0pt plus .5pt}

\newcommand{\ts}{\hspace*{0.4em}}
\def\onedot{.,\ }

\def\ie{i.e\onedot} 
 
\def\etal{et al.\ }

\title{\LARGE Contrastive Learning for Self-Supervised Pre-Training\\ of Point Cloud Segmentation Networks With Image Data\vspace{-4mm}}
\author{
Andrej Janda,\ts Brandon Wagstaff,\ts Edwin G.\ Ng,\ts and Jonathan Kelly \\
Space \& Terrestrial Autonomous Robotic Systems Laboratory, University of Toronto, Canada\\
  \texttt{<first\_name>.<last\_name>@robotics.utias.utoronto.ca}}

\begin{document}
\maketitle

\begin{abstract}
  Reducing the quantity of annotations required for supervised training is vital when labels are scarce and costly.
  This reduction is particularly important for semantic segmentation tasks involving 3D datasets, which are often significantly smaller and more challenging to annotate than their image-based counterparts.
  Self-supervised pre-training on unlabelled data is one way to reduce the amount of manual annotations needed.
  Previous work has focused on pre-training with point clouds exclusively. While useful, this approach often requires two or more registered views.
  In the present work, we combine image and point cloud modalities by first learning self-supervised image features and then using these features to train a 3D model.
  By incorporating image data, which is often included in many 3D datasets, our pre-training method only requires a single scan of a scene and can be applied to cases where localization information is unavailable.
  We demonstrate that our pre-training approach, despite using single scans, achieves comparable performance to other multi-scan, point cloud-only methods.
\end{abstract}

\section{Introduction}
\label{sec:introduction}

The two most common representations used for robotic scene understanding tasks are images and point clouds.
Images are dense and feature-rich, but their lack of depth information limits how well they are able to model 3D environments when used alone.
Although point clouds circumvent many of the limitations inherent to images, they are notoriously hard to annotate.
This annotation difficulty is a key limiting factor for many state-of-the-art data-driven scene understanding algorithms that require large, annotated datasets \cite{he2017mask,charles2017PointNet,jiang2020pointgroup,choy20194d}.
Generating labels requires human annotators to manipulate the clouds by zooming, panning, and rotating to select points of interest.
Annotators then have to separate points that belong to a particular object from the background and other occluded points.

\begin{figure}
  \centering
  \vspace{2mm}
  \includegraphics[width=0.96\columnwidth]{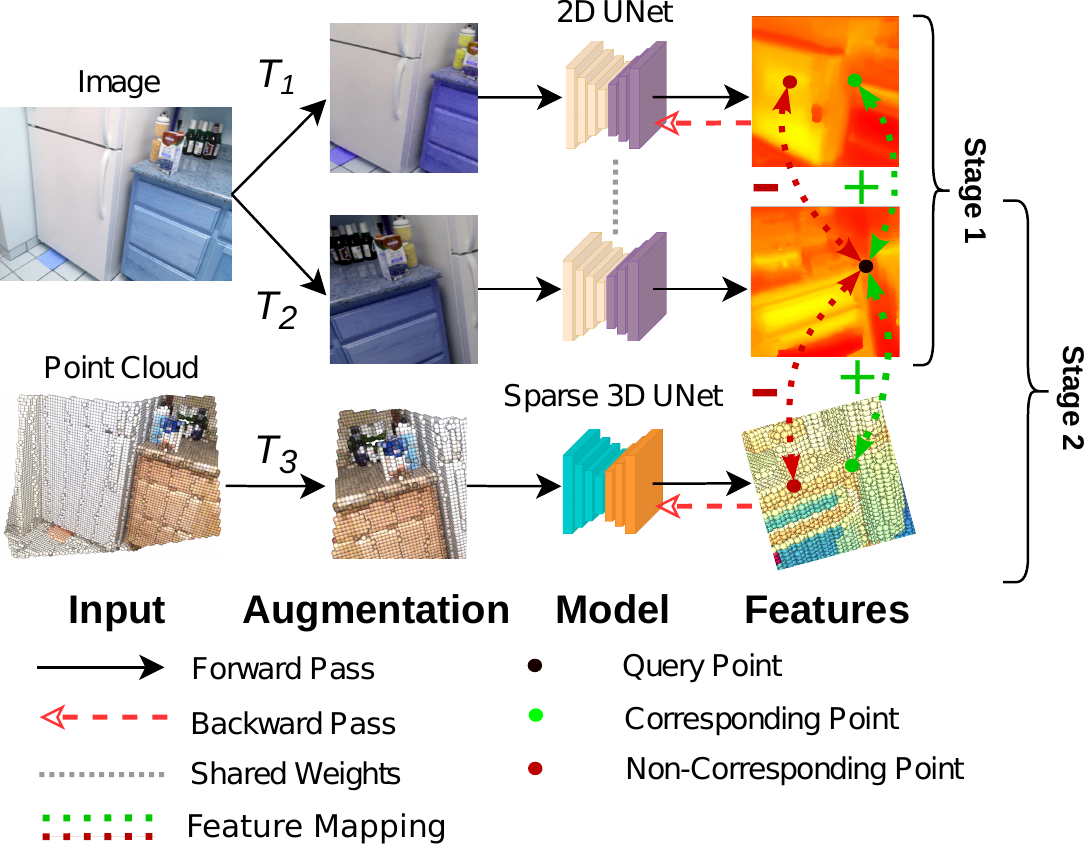}
  \vspace{1mm}
  \caption{Overview of our multimodal contrastive learning framework. The first stage pre-trains a 2D model from image pixel data. The second stage uses the features from the 2D model as targets to pre-train a 3D model.}
  \label{fig:overview}
  \vspace{-5mm}
\end{figure}

The difficulty of annotating point clouds has resulted in considerable effort and labelling times for existing datasets.
For example, the SemanticKITTI dataset \cite{behley2019semantic}, which has 518 square tiles of 100 metres length each, required 1,700 hours to label.
ScanNet \cite{dai2017scannet}, which has 1,600 reconstructed scenes of indoor rooms, took about 600 hours to label \cite{zhang2021self}.
Despite the substantial labelling time, 3D datasets are still significantly
smaller than comparable image-only datasets.

The labelling effort required for 3D data is the reason we seek, herein, to reduce the volume of annotations necessary.
Previous work has demonstrated that \emph{self-supervised contrastive pre-training} is an effective approach for improving performance on scene understanding tasks with raw, unlabelled point cloud data \cite{xie2020pointcontrast, hou2021Exploring, zhang2021self}.
A key limitation of existing 3D pre-training methods is that they neglect the information-rich images that are often available as part of 3D datasets.
We propose a pre-training method that leverages images as an additional modality, by learning self-supervised image features that can be used to pre-train a 3D model.
Our learning method is split into two stages.
The first stage (Stage 1) learns image features via a self-supervised contrastive learning framework.
The second stage (Stage 2) applies the same contrastive learning framework to pre-train a 3D model, making use of the 2D features learned in Stage 1.
By incorporating visual data into the pre-training pipeline, we obtain a notable advantage: only a single point cloud scan and the corresponding image are required during pre-training.
The use of a single scan obviates the need for two or more overlapping 3D views, which are required by many point cloud-only approaches.
Notably, the use of a single scan improves the scalability of our approach, since we require raw 3D data only, as opposed to multiple scans that have been aggregated using a robust mapping pipeline for data association.

Through extensive experimentation, we compare our pre-training approach with existing point cloud-only approaches on several downstream tasks and across several datasets.
We find that our method performs competitively with methods that use multiple overlapping point cloud scans, despite having access to single scans and images only.
In short, we make the following contributions:
\begin{itemize}
  \item we describe a self-supervised method for extracting visual features from images and using them as labels to pre-train 3D models via a contrastive loss;
  \item we provide visualizations demonstrating that the features learned capture structure, such as lines and surface patches, in the input image and point cloud;
  \item we demonstrate that a model trained using features learned from raw images improves performance on 3D segmentation and object detection tasks.
\end{itemize}

\section{Related Work}
\label{sec:related-work}

Our approach builds upon existing work on self-supervised contrastive learning with images and point clouds.
These techniques typically produce two augmented versions of each input sample by applying a series of transformations with different sets of parameters.
Any two augmented samples derived from the same initial input are referred to as a \emph{positive} pair, while those not derived from the same input are referred to as a \emph{negative pair}.
Subsequently, a contrastive loss minimizes the (Euclidean) distance between model outputs for positive pairs, while maximizing the distance between model outputs for negative pairs.
Here, we provide a brief summary of algorithms that use this approach to pre-train image and point cloud networks.

\subsection{Contrastive Learning in 2D}
Self-supervised pre-training using image features has proven to be a successful approach for downstream image-classification tasks, achieving comparable performance to supervised pre-training, as demonstrated by SimCLR \cite{chen2020simple}.
Self-supervised approaches require large batches of negative samples, however, which are not always feasible to obtain.
An alternative is to store previous feature encodings in a memory bank so that they do not have to be recomputed \cite{wu2018unsupervied}.
Momentum Encoders (MoCo) \cite{he2020momentum} extends memory banks with negative samples computed from a separate encoder that is updated according to a moving average of the current model parameters.
BYOL \cite{grill2020bootstrap} uses the same dual-encoder architecture as MoCo, but does not include a memory bank or use any negative samples.
SimSiam \cite{chen2021exploring} modifies BYOL by sharing weights between the two encoders instead of using a moving average.
Clustering methods have also proven effective at selecting negative samples that are the most informative \cite{caron2020unsupervised,li2021prototypical}.
Similarly, SupCon \cite{khosla2021supervised} uses pseudo-labels, generated from a partially trained model, to prevent a negative sample of the same class as the query point from being drawn.
Negative mining techniques \cite{chuang2020debiased, robinson2021contrastive, kalantidis2020hard} also seek to improve performance through negative sample selection.
Notably, for segmentation tasks, self-supervised pre-training of pixel-level features \cite{wang2021dense,xie2021propagate,wang2021exploring} offers improved downstream performance when compared to image-level features.

\subsection{Contrastive Learning in 3D}
The architectures used for pre-training of image networks can be adapted to work with point cloud networks.
PointContrast \cite{xie2020pointcontrast} is an early example that augments two overlapping 3D scans with random rotations and colour transformations.
Known, corresponding points between the two transformed scans form a positive pair, while all other points are considered as negative samples in the contrastive loss function.
Contrastive Scene Contexts (CSC) \cite{hou2021Exploring} extends PointContrast with an additional partitioning scheme.
DepthContrast \cite{zhang2021self} augments a single scan and learns feature vectors at the scan-level instead of at the point-level.
Jiang \etal extend SupCon to point clouds in \cite{jiang2021Guided}.
SegContrast \cite{nunes2022segcontrast} uses a hand-engineered (\ie non-learning-based) clustering algorithm to segment a scene prior to positive and negative sample selection.

\subsection{Multimodal Contrastive Learning}
By projecting 3D points into images, algorithms such as Pri3D \cite{hou2021pri3d} and SimIPU \cite{zhenyu2022simipu} leverage 3D data when pre-training models for downstream 2D scene understanding tasks.
In \cite{hou2021pri3d} and \cite{zhenyu2022simipu}, the pixel-point pairs that map to the same physical 3D location are used as positive pairs in a (pixel-only) contrastive loss.
Alternatively, pre-training with image data can improve downstream performance on 3D scene understanding tasks.
CrossPoint \cite{afham2022crosspoint} applies a contrastive learning objective to global scene features generated from synthetic point clouds of computer-modelled objects and the corresponding rendered images.
A major limitation of CrossPoint is that it operates on synthetic object-centric datasets and has not been shown to scale effectively to real-world 3D scans.
P4Contrast \cite{yunze2020p4contrast} performs sensor fusion and self-supervised pre-training on combined 2D-3D inputs, both at training time and runtime.
Pixel-to-Point Knowledge Transfer \cite{liu2021learning} learns point-level features from pre-trained pixel-level features.
This approach is most similar to our method, with the differences being the specific architecture and the use of images from the desired, target dataset.

\section{Methodology}
\label{sec:method}

In this section, we provide an overview of the formulation of our contrastive learning framework.
We describe the two distinct and sequential stages of our framework, shown in \Cref{fig:overview}.
The first stage applies a 2D CNN to generate image features at the pixel level, based on a contrastive loss on the individual pixels.
The second stage then uses these image features to train a 3D model.

\subsection{Self-Supervised Contrastive learning}
\label{sec:method:background}

Contrastive learning aims to produce features that are distinguishable between unique inputs.
Following the formulation in \cite{lekhac2020contrastive}, a query point $\mathbf{x}_i \in \mathbb{R}^{N}$ is first sampled from the dataset.
The query point is then augmented by sequentially applying one or more individual transformations $t: \mathbb{R}^{N} \rightarrow \mathbb{R}^{N}$, forming a composite function, $T(\mathbf{x}_i) = (t_1 \circ \dots \circ t_S)(\mathbf{x}_i)$, where all transformation parameters are sampled randomly.
Applying two separate transformations to the query point results in the positive pair $(T(\mathbf{x}_i), T^{+}(\mathbf{x}_i))$.
Negative samples are selected as those data points or augmentations not derived from the query point.

All points (positive and negative) are subsequently fed through an encoder $f_{\boldsymbol{\theta}}$ to obtain a feature $\mathbf{v} = f_{\boldsymbol{\theta}}(\mathbf{x})$, where $\mathbf{v} \in \mathbb{R}^D$.
This encoder forms the backbone of the model that we are trying to initialize.
The feature is then passed through a decoder $\mathbf{z} = h_{\boldsymbol{\phi}}(\mathbf{v})$, where $\mathbf{z} \in \mathbb{R}^M$ and $M \leq D$.
If the features are desired directly then the decoder can simply apply the identity transform.
The decoder outputs are normalized such that $\lVert \mathbf{z} \lVert^{2} = 1$ to improve the stability of the gradient updates during training.
Once pre-trained, only the encoder parameters $\boldsymbol{\theta}$ are retained as part of the initialized backbone; the decoder parameters $\boldsymbol{\phi}$ are discarded.

The most common and successful contrastive objective function is the InfoNCE (Info Noise Contrastive Estimation) loss function \cite{oord2018representation}.
The loss is defined over the set of query points as%
\vspace{-2mm}
\begin{equation}
  \mathcal{L} = -\sum_{i = 1}^{N}\log \frac{\exp(\mathbf{z}_{i} \cdot \mathbf{z}^{+}_{i} / \tau)}{\exp(\mathbf{z}_{i} \cdot \mathbf{z}^{+}_{i} / \tau) + \sum^{K}_{j}\exp(\mathbf{z}_{i} \cdot \mathbf{z}^{-}_{j} / \tau)},
  \label{eq:contrastive_loss}
\end{equation}
where $\tau \in (0,1]$ is a temperature parameter that controls the smoothness of the latent (encoded) representations and $K$ is a hyperparameter that determines the number of negative features to sample.
For simplicity, it is common to set $K = N$ and to take all negative samples as query points.
The similarity between features is computed as the dot product, although other suitable distance functions exist.

\subsection{Stage 1 -- Image Features}
\label{sec:method:stage1}
We utilize the ResUNet architecture from Godard at al. \cite{godard2019Digging} to extract 2D pixel-level features and modify the decoder to compute a 16-dimensional feature vector for each pixel in the input image.
We pre-load the weights of the encoder from a model trained on the large ImageNet corpus \cite{deng2009imagenet}.
To pre-train the full model, images are selected from a desired pre-training dataset.
We follow roughly the same data augmentation strategy and use the same InfoNCE loss function as SimCLR \cite{chen2020simple}, except that we compare pixel-level features instead of image-level features.
Pixels that map back to the same coordinates in the original image are considered as positive samples, while all others (including those from other images in a batch) are considered as negative samples.

\subsection{Stage 2 -- Point Features}
\label{sec:method:stage2}
In this stage, we pre-train a 3D point-level feature extraction model using the pixel-level features from Stage 1.
We apply the 3D model from \cite{xie2020pointcontrast} and treat the final $1\times1$ convolution as the decoder, which we initialize from scratch for training on downstream tasks.
Each point cloud is also augmented so that the model learns to be invariant to differences in orientation, point density, and colour fluctuations.
The 2D network is held frozen and the fixed 2D features act as a target for the 3D model to learn.
Mapping between 3D points and 2D features is done via perspective projection.
Each pixel-point match forms a positive pair $(\mathbf{z}_i,\mathbf{z}^+_i)$, where $\mathbf{z}_i$ and $\mathbf{z}^+_i$ represent the feature vectors of a 3D point and the corresponding pixel, respectively.
The feature vector of any other point is considered a negative sample and represented as $\mathbf{z}^-_j$.
We use the InfoNCE loss (see \Cref{eq:contrastive_loss}) to train the 3D model.
The final parameters of the 3D model are then used on downstream 3D scene understanding tasks.

\begin{figure*}
  \begin{subfigure}[t]{0.29\textwidth}
    \centering
    \includegraphics[width=4.6cm]{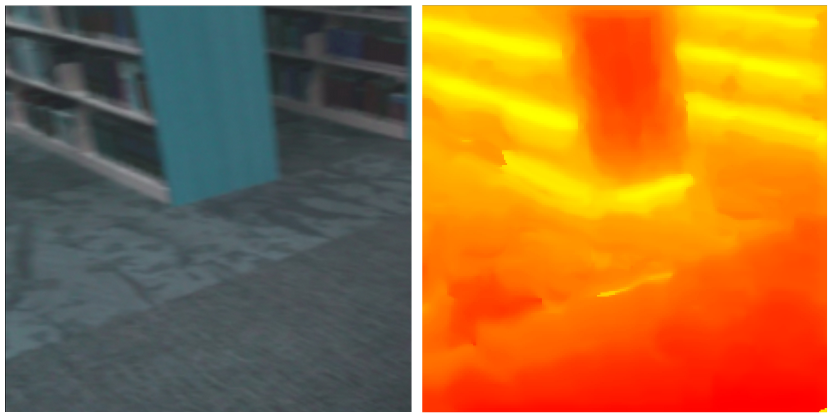}
    \includegraphics[width=4.6cm]{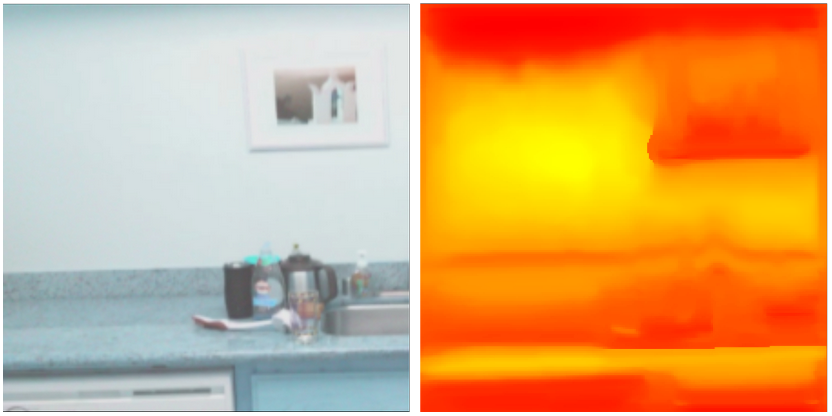}
    \caption{}
    \label{fig:features2dvis}
  \end{subfigure}
  \hfill
  \begin{subfigure}[t]{0.29\textwidth}
    \centering
    \includegraphics[width=4.8cm]{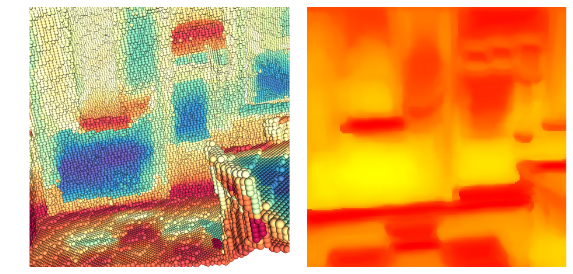}
    \includegraphics[width=4.8cm]{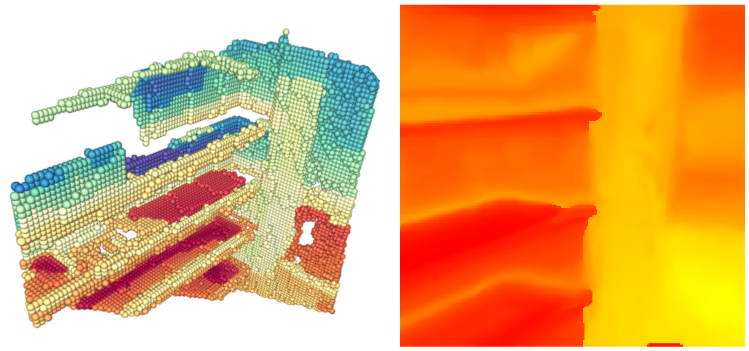}
    \caption{}
    \label{fig:features2d-3dvis}
  \end{subfigure}
  \hspace{0.2cm}
  \begin{subfigure}[t]{0.37\textwidth}
    \centering
    \includegraphics[width=5.8cm]{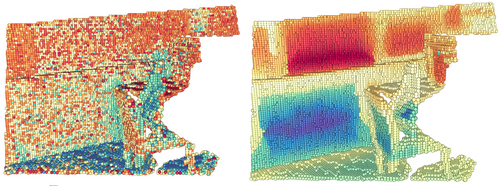}
    \includegraphics[width=5.8cm]{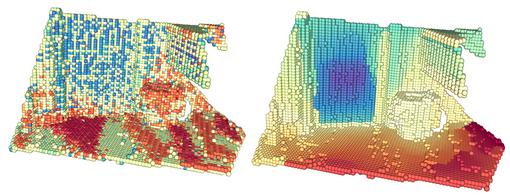}
    \caption{}
    \label{fig:FeaturesScratchVsPretrained}
  \end{subfigure}
  \caption{Visualization of 2D and 3D feature vectors. The columns in each figure, from left to right, are: (a) input images and corresponding pre-trained 2D features; (b) pre-trained point (cloud) features and corresponding pre-trained 2D features; and (c) point features before and after pre-training using our method.}
  \label{fig:three_graphs}
  \vspace{-4mm}
\end{figure*}

\subsection{Training and Evaluation Details}

The pre-training pipeline requires that the image-to-point cloud mapping is known and that the camera intrinsic parameters are available.
The extrinsic pose of the camera can be found using a variety of localization techniques or may be determined directly (if stereo or an RGB-D sensor is used).
When an entire scene is viewed by a monocular camera from multiple poses, there is a risk of incorrect mappings due to occlusions (i.e., points being mapped to pixels even if the points would not actually be visible).
We use datasets without occlusions only, avoiding this issue.
Also, notably, occlusions can be removed using methods such as \cite{katz2007Direct}.
The removal of occlusions allows training to leverage several views of the same point cloud, which further increases the training set size.

Once pre-training has been completed, downstream training only requires point clouds to fine-tune the 3D model.
Evaluation is then performed on that model itself.

\section{Experiments}
\label{sec:results}

In this section, we benchmark the performance of our pre-trained model on a variety of popular datasets and tasks.
We compare to state-of-the-art baselines on three different downstream tasks: semantic segmentation, instance segmentation, and object detection.
\Cref{tab:downstreamResults} provides an overview of our results, across different datasets and tasks.

\subsection{Datasets}
\label{sec:results:datasets}

We utilize three indoor stereo datasets that are common benchmarks for 3D contrastive learning \cite{xie2020pointcontrast, hou2021Exploring, zhang2021self, jiang2021Guided}, together with an extra outdoor lidar dataset \cite{behley2019semantic}.
We use ScanNet to pre-train our backbone network as it is the largest indoor dataset available.
ScanNet shares many of the same classes and types of indoor scenes as S3DIS and SUNRBGD and therefore is a good candidate for demonstrating the impact of pre-training on a large unlabelled dataset.
We also use ScanNet to pre-train our SemanticKITTI model, in order to evaluate how learned 3D features can generalize to substantially different environments and sensor types.

ScanNet \cite{dai2017scannet} is comprised of roughly 1,600 reconstructed scenes of indoor environments.
The raw data include individual RGB-D images.
ScanNet scenes are mostly of individual rooms that range in size from 2 metres to 10 metres on a side, with a standard ceiling height of about 3 metres.
The rooms contain 20 different classes of labelled objects.
We use the dataset-defined training and validation splits, and use the validation set as our test set.

S3DIS \cite{armeni20163D} is a much smaller dataset and is comprised of only 300 reconstructed scenes.
However, the size of these scenes varies quite drastically; some are of rooms and others are of entire auditoriums.
The scenes are mainly of indoor office environments and each scan is an RGB-D image.
There are 13 different semantic classes. All scenes contain point-level semantic and instance labels.
Following \cite{xie2020pointcontrast, hou2021Exploring}, we use the Area 5 split for  validation and testing.

SunRGBD \cite{song2015sunrgbd, janoch2011category, xiao2013database, silberman2012indoor} is a 3D dataset of indoor office environments.
SunRGBD contains roughly 10,000 RGB-D scans and corresponding 3D bounding box annotations for 10 different classes, with labels similar to those of ScanNet and S3DIS.
There is no scene reconstruction available, unfortunately.
We use the dataset-defined training and validation split.

SemanticKITTI \cite{behley2019semantic, geiger2012are} contains roughly 25,000 laser scans of outdoor driving environments.
It has 20 different classes with 3D semantic and instance labels.
The scans cover 360$^\circ$ around the vehicle, going out to a range of about 20 metres, with the spatial resolution decreasing with distance.
We use sequences 1--7 and 9--10 as our training set and sequence 8 as our validation set.

\begin{table*}[t]
  \centering
  \resizebox{\textwidth}{!}{
    \begin{threeparttable}
      \begin{tabular}{ c c | c c | c c c | c | c}
        \toprule
        \multicolumn{2}{c|}{\multirow{2}{*}{\textbf{Pre-Training Method}}} & \multicolumn{2}{c|}{\textbf{S3DIS}} & \multicolumn{3}{c|}{\textbf{ScanNet}} & \textbf{KITTI}       & \textbf{SUNRGBD}                                                                                                 \\
                                                                           &                                     & Semantic                              & Instance             & Semantic             & Instance             & Object               & Semantic             & Object               \\
        \midrule
                                                                           & Scratch                             & 65.1                                  & 53.0                 & 67.4                 & 49.0                 & 35.2                 & 41.0                 & 32.0                 \\
                                                                           & Supervised                          & 70.2 (+5.1)                           & 56.2 (+3.2)          & --                   & --                   & --                   & --                   & --                   \\

        \midrule
        \multirow{2}{*}{\textbf{Multi-Scan}}                               & PointContrast                       & 66.2 (+1.1)                           & 54.8 (+1.8)          & 66.9 (-0.5)          & 49.1 (+0.1)          & \textbf{36.7 (+1.5)} & 42.1 (+1.1)          & 34.2 (+2.2)          \\
                                                                           & CSC                                 & \textbf{69.0 (+3.9)}                  & \textbf{57.8 (+4.8)} & \textbf{67.6 (+0.2)} & \textbf{49.3 (+0.3)} & 36.1 (+0.9)          & \textbf{43.0 (+2.0)} & \textbf{35.1 (+3.1)} \\
        \midrule
        \multirow{2}{*}{\textbf{Single-Scan}}                              & DepthContrast                       & 64.9 (-0.2)                           & 52.3 (-0.7)          & 67.4 (+0.0)          & \textbf{48.7 (-0.3)} & 33.9 (-1.3)          & 42.0 (+1.0)          & 32.9 (+0.9)          \\

                                                                           & \textbf{Ours}                       & \textbf{66.5 (+1.4)}                  & \textbf{55.8 (+2.8)} & \textbf{67.7 (+0.3)} & 48.5 (-0.5)          & \textbf{37.7 (+2.5)} & 42.0 (+1.0)          & \textbf{33.1 (+1.1)} \\
        \bottomrule
      \end{tabular}
    \end{threeparttable}
  }
  \vspace{1mm}
  \caption{Downstream performance comparison of pre-training methods. Semantic segmentation uses the mIOU metric, while both instance segmentation and object detection tasks use the mAP@0.5 metric with a minimum correct overlap ratio of 0.5. The best performance on each task and dataset for the single- and multi-view categories are highlighted in bold.}
  \label{tab:downstreamResults}
  \vspace{-3mm}
\end{table*}

\subsection{Baselines}
\label{sec:results:baselines}

To verify the effectiveness of our method, we compare against three state-of-the-art baseline algorithms: PointContrast \cite{xie2020pointcontrast}, Contrastive Scene Contexts (CSC) \cite{hou2021Exploring}, and DepthContrast \cite{zhang2021self}.
Both PointContrast and Contrastive Scene Contexts require pairs of scans with known poses and at least 30\% overlap, while DepthContrast and our method operate on single scans only.

Due to our own resource limitations, we run DepthContrast with a batch size of 32 on a single graphics processing unit (GPU), instead of with a batch size of 1,024 split across 32 GPUs.
We run DepthContrast for 40 epochs instead of 400, which still takes twice as long as any other method.
This should serve as a more fair comparison between algorithms when access to large compute clusters is not possible.
Where applicable, we also compare against a fully-supervised backbone to give an idea of a reasonable upper-bound on performance improvement.

\subsection{Implementation Details}
\label{sec:results:implementation}

The backbone network we use is the same as that in \cite{xie2020pointcontrast, hou2021Exploring}, with the sparse convolutional library developed by Choi et al.\ \cite{choy20194d}.
The backbone has a Res16UNet34 structure with non-bottleneck blocks and a maximum feature embedding size of 256.
The outputs of the model are directly used for semantic segmentation.
Instance segmentation is performed with the same backbone to extract features, which are subsequently fed into PointGroup \cite{jiang2020pointgroup}.
Object detection is performed with VoteNet \cite{qi2019deep}, using the sparse voxel backbone instead of PointNet++.

For pre-training of the 2D backbone on images, we use a batch size of 64 image pairs.
For each pair, we sample 4,092 pixels to contrast in our loss function.
The total loss for each training iteration is the sum of the loss for each image pair.
The loss function uses a $\tau$ of 0.4, a learning rate of 0.01, an SGD optimizer with a momentum of 0.9, dampening of 0.1, and a weight decay of 0.004.
We decrease the learning rate according to an exponential scheduler with an exponential rate of 0.99.
Positive samples are generated using the image transformations from \cite{chen2020simple}.
We pre-train the image network for 20,000 iterations.

Pre-training of the 3D backbone is carried out on the ScanNet dataset.
We use a batch size of eight scan-image pairs and select 2,000 point-pixel correspondences per pair.
This is a significantly smaller number than for 2D pre-training because of the increased compute and memory usage required by point cloud data (compared to image-data).
The images are centre-cropped to $224 \times 224$ pixels to conform to the expected input size of the 2D model.
We voxelize the points, with a voxel size of 5 cm per side.
The hyper-parameters are the same as those for 2D training except that the learning rate is set to 0.1.
The 3D model is pre-trained for 20,000 iterations.

All downstream tasks use the pipeline and parameters from CSC \cite{hou2021Exploring}.
Note that we obtain slightly different numbers from those reported in \cite{hou2021Exploring}, possibly due to a required update to the core sparse convolution library for our newer GPU.
We also use a single GPU instead of eight.
In this case, all pre-training is run from scratch.

\subsection{Feature Visualization}
\label{sec:results:featureVisualization}

We first verify that the 2D features learned using our 2D pre-training scheme have some connection to the original image by following the approach from \cite{choy2019Fully}.
We bring each pixel feature into a 1D color space using the t-SNE \cite{maaten2008Visualizing} algorithm.
A heatmap is then applied and the features are composited to yield an image using each pixel's original coordinates.
\Cref{fig:features2dvis} shows a comparison between our visualizations and the original image.
These visualizations indicate a clear relationship between input image and output features.
The heatmaps tend to highlight structures such as lines and surface patches that are present in the input image.
Key parts of the image share similar feature vectors (e.g., individual shelves, tables and chairs).
The similarity of features within classes and their parts should make extracting the class label of a point easier for the decoder.

\Cref{fig:features2d-3dvis} shows the relationship between the 2D and 3D features.
There is a clear mapping between the image and the corresponding point cloud heatmaps.
This correlation qualitatively verifies that the 3D model has indeed learned to `mimic' the features of the 2D model without relying on the image.
Parts of the scene, such as walls and shelves, have consistent features that are present across both modalities.
\Cref{fig:FeaturesScratchVsPretrained} shows the difference between randomly initialized and pre-trained features.
Features that are pre-trained follow visible object boundaries.

\subsection{Semantic Segmentation}
\label{sec:results:semantic}

The effects of several pre-training methods (including our own) on downstream tasks are detailed in \Cref{tab:downstreamResults}.
Fully-supervised pre-training on ScanNet has a drastic impact on final, downstream performance ($+5.1$ mIOU) and is used as a rough upper bound on expected performance.
Almost all pre-training algorithms improve downstream results, except DepthContrast.
Our algorithm's performance is comparable to PointContrast: both methods achieve an mIOU increase of more than 1\% but perform worse than CSC.

\begin{figure*}
  \begin{subfigure}[t]{0.3\textwidth}
    \centering
    \includegraphics[width=5.4cm]{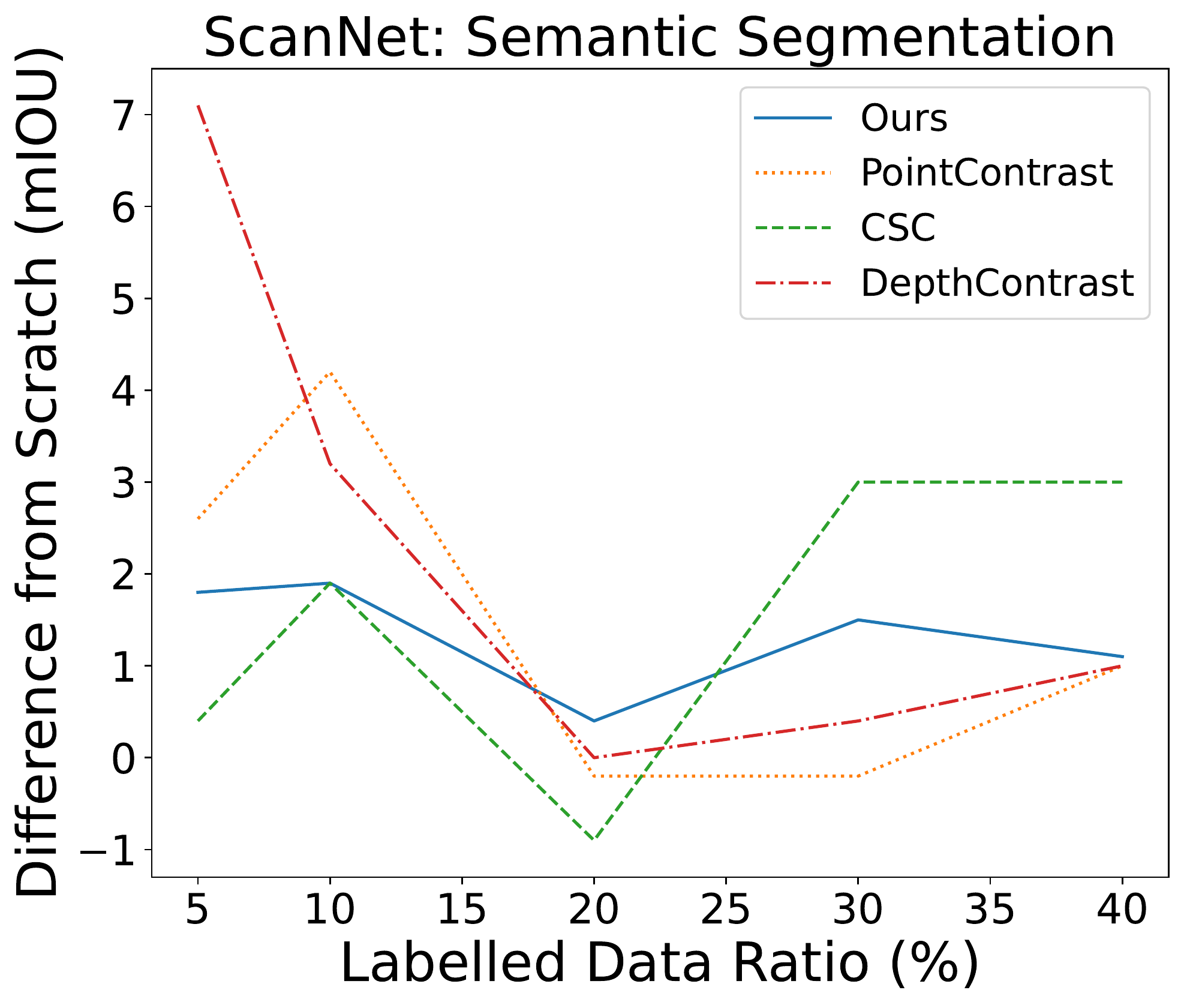}
    \caption{}
    \label{fig:ScannetVaryingSemantic}
  \end{subfigure}
  \hfill
  \begin{subfigure}[t]{0.3\textwidth}
    \centering
    \includegraphics[width=5.4cm]{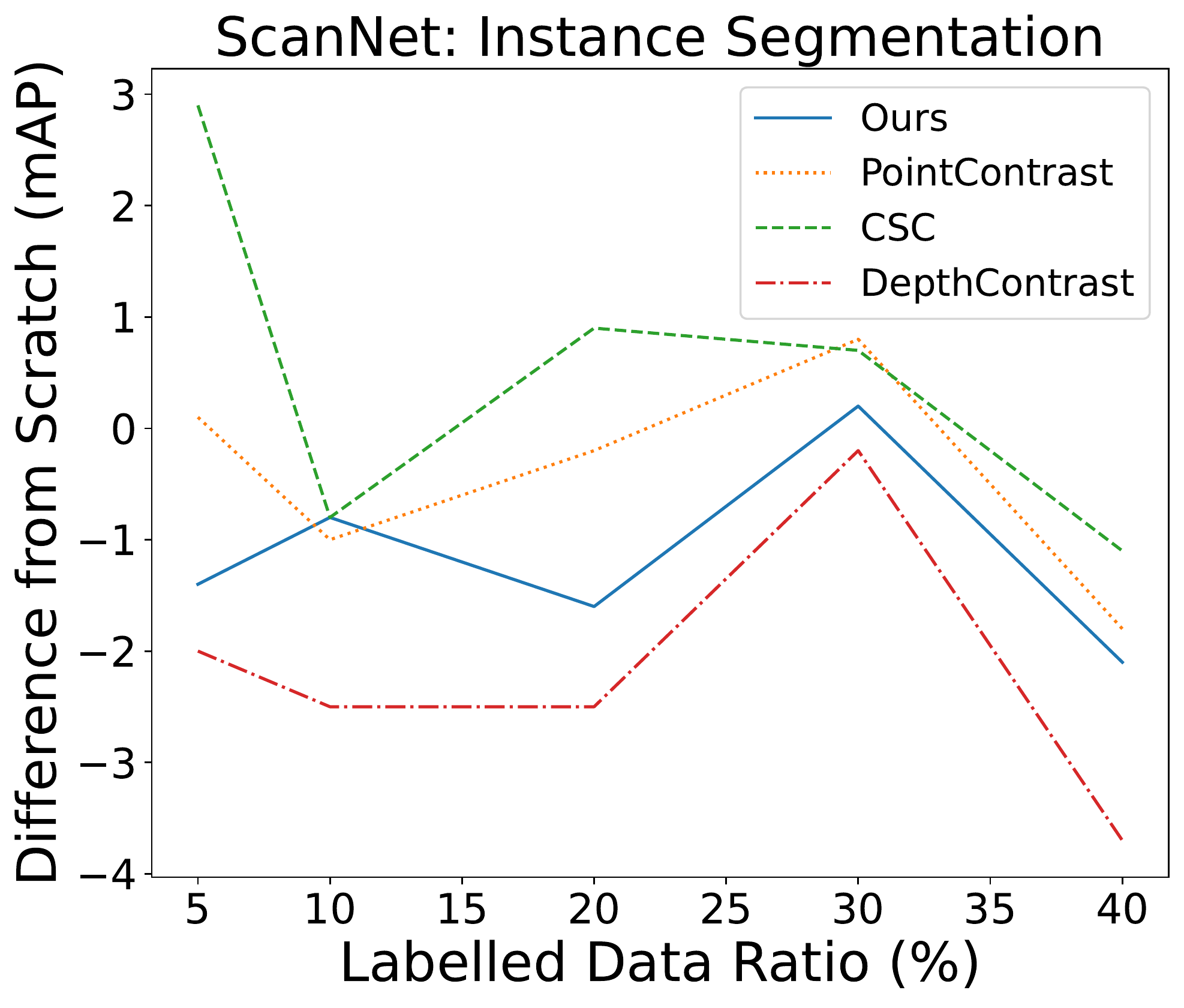}
    \caption{}
    \label{fig:ScannetVaryingInstance}
  \end{subfigure}
  \hfill
  \begin{subfigure}[t]{0.3\textwidth}
    \centering
    \includegraphics[width=5.4cm]{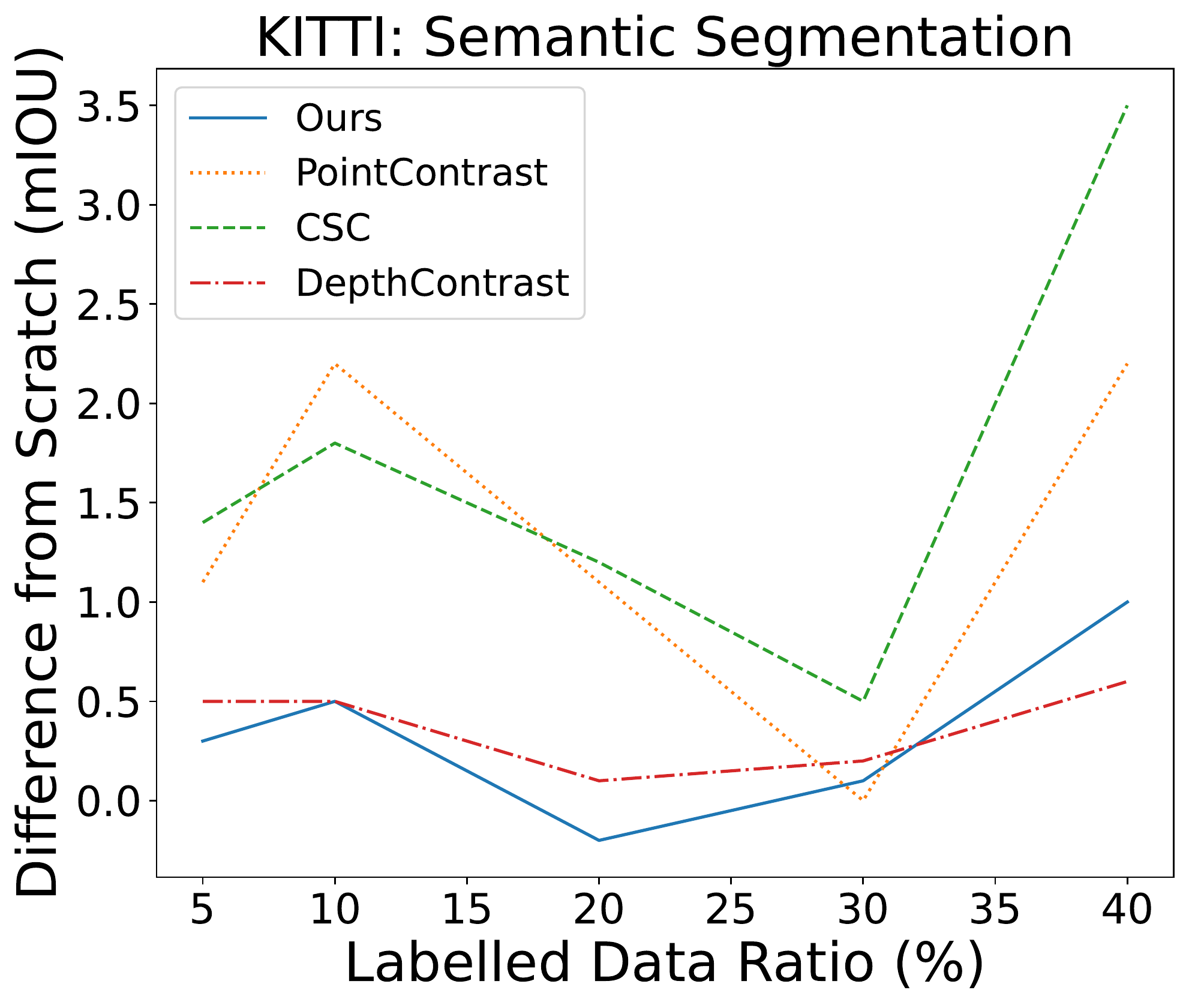}
    \caption{}
    \label{fig:KittiVaryingSemantic}
  \end{subfigure}
  \caption{Effects of varying training data proportion on the performance boost of pre-trained models.}
  \label{fig:varyingDataProportions}
  \vspace{-4mm}
\end{figure*}

When ScanNet is used for both pre-training and supervised learning, all methods fail to improve semantic segmentation.
Although we see no improvement on ScanNet, we find that pre-training methods do improve performance when the number of annotations during supervised training on downstream tasks is reduced, as seen in \Cref{fig:ScannetVaryingSemantic}.
Using varying proportions of labels simulates the use case where only a portion of the data collected are annotated.

Pre-training on ScanNet shows performance improvements on SemanticKITTI as well, even though the datasets are quite different.
In \Cref{fig:KittiVaryingSemantic}, we test the effect of varying the amount of labelled data on SemanticKITTI.
We find that in almost all scenarios, pre-training methods help.
The performance improvements on S3DIS and SemanticKITTI suggest that exposing models to different environments and sensor setups assists in learning features that are better able to generalize.

\subsection{Instance Segmentation}
\label{sec:results:instance}

As in the previous section, we evaluate the performance of pre-training on ScanNet and training on S3DIS for instance segmentation.
Results are shown in \Cref{tab:downstreamResults}.
Almost all pre-training methods are able to come close to or even to surpass supervised results, with the exception (again) being DepthContrast.
The best-performing method is CSC, which achieves a performance boost of 4.8 mAP@0.5.
Our own method performs similarly to supervised pre-training, with a boost of 2.8.

As with semantic segmentation, all methods struggle to improve on ScanNet instance segmentation, even when access to labelled data is limited as shown in \Cref{fig:ScannetVaryingInstance}.
In most cases, the algorithms perform substantially worse than training from scratch.
This is in spite of the boost to semantic segmentation with the same data ratios.

\subsection{Object Detection}
\label{sec:results:object}

Our last test case is the downstream task of object detection.
Results for the ScanNet and SunRBGD datasets are found in \Cref{tab:downstreamResults}.
This task is interesting because, unlike segmentation, it operates at an object level instead of at the point level (which is our pre-training objective).
All methods besides DepthContrast show a strong improvement on both datasets.
CSC exhibits the best performance with a boost on SunRGBD of 3.1 mAP, while our method attains the largest boost on ScanNet with an increase of 2.5 mAP.
For the segmentation task, using the same dataset for both pre-training and the downstream task yields no performance gain.
However, object detection on ScanNet shows strong improvements even when the full dataset is available.
This outcome suggests that pre-trained point-level features generalize well to object-level tasks when using the same raw data for both.

\section{Analysis}
\label{sec:analysis}

\subsection{Pre-Training on Out-of-Distribution Data}
\label{sec:results:out-of-distribution}

We find that using a pre-training dataset that is out-of-distribution relative to the downstream dataset leads to more consistent and greater performance gains than using the same dataset for both pre-training and fine-tuning.
\Cref{tab:downstreamResults} shows that using model weights pre-trained on ScanNet offers very little performance improvement when fine-tuned on ScanNet.
However, those same weights result in much larger gains when applied to a different dataset such as S3DIS or SemanticKITTI.
A similar discrepancy is found when varying the proportion of labelled data available.
Using weights pre-trained and fine-tuned on ScanNet, we find that the biggest performance gain occurs when the proportion of labelled data available during fine-tuning is the most restricted (shown in \Cref{fig:ScannetVaryingSemantic,fig:ScannetVaryingInstance}).
The same weights have the opposite effect when fine-tuning on SemanticKITTI, which sees the largest performance boost as the proportion of labelled data increases, (shown in \Cref{fig:KittiVaryingSemantic}).
The discrepancy suggests that the relationship between the data available during pre-training and the data available during fine-tuning is a key contributing factor to the effectiveness of pre-training.

\subsection{Inconsistencies in Performance Boost}
\label{sec:results:inconsistencies}

\begin{figure}[t!]
  \centering
  \includegraphics[trim=0 0 0 20pt,clip,width=8.25cm]{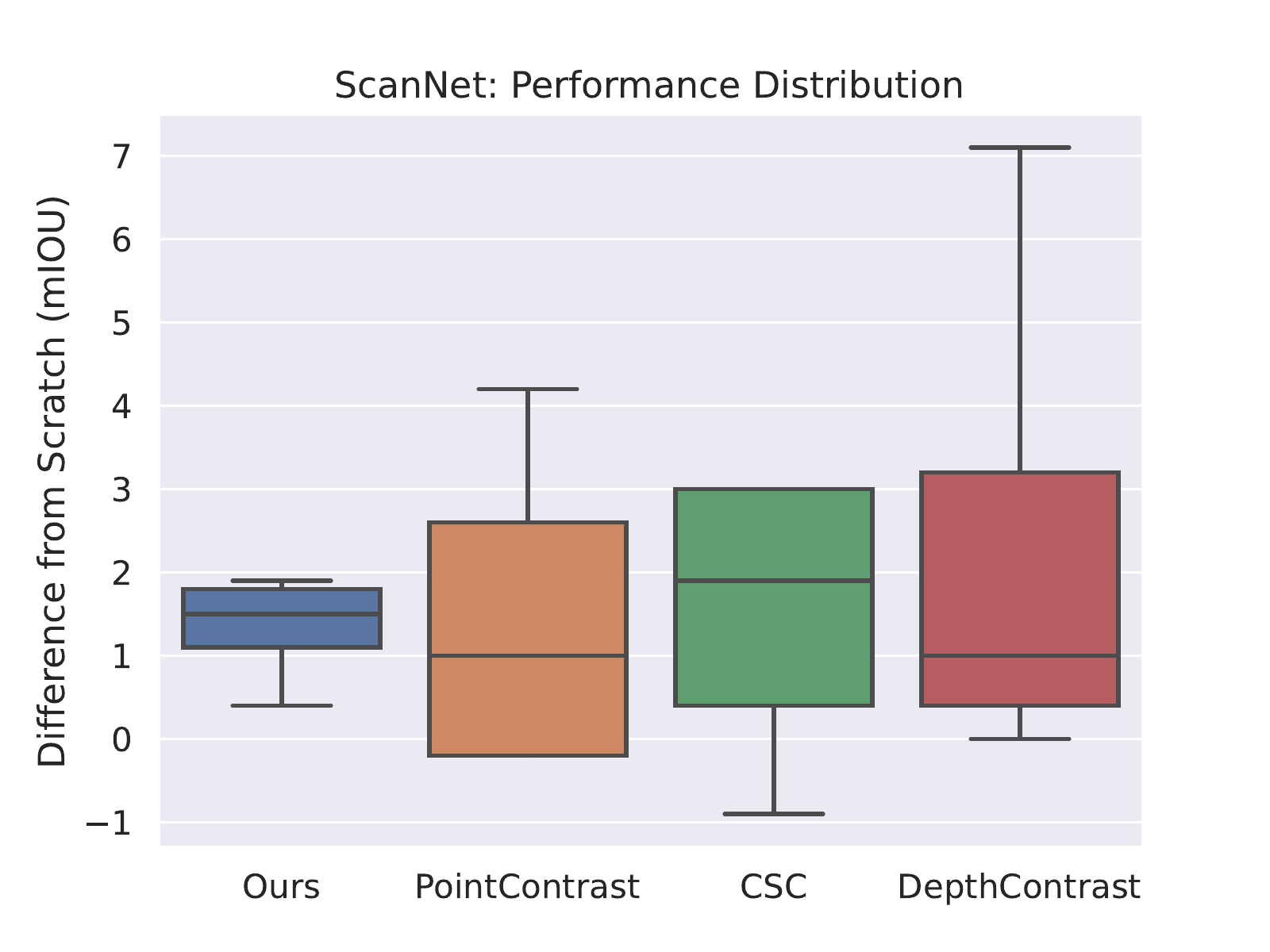}
  \caption{Distribution of performance differences with respect to scratch performance on ScanNet semantic segmentation across labelled data ratios used in \Cref{fig:ScannetVaryingSemantic}.}
  \label{fig:distribution}
  \vspace{-4mm}
\end{figure}

In this section, we investigate the relationship between unlabelled and labelled data proportions on downstream performance, by limiting the available label amounts (details shown in \Cref{fig:ScannetVaryingSemantic,fig:ScannetVaryingInstance,fig:KittiVaryingSemantic}).
While we expect the benefits of pre-training to increase during downstream supervised training in situations when fewer labels are available, we find that the actual performance gains are inconsistent.
To demonstrate this inconsistency, we plot the distribution of semantic segmentation accuracy with varying amounts of labelled ScanNet data in \Cref{fig:distribution}.
The mean improvement of our method is higher than that of both PointContrast and DepthContrast.
Our method also has a much smaller variance and greater lower bound on the performance gain.
Therefore, given a dataset with a limited amount of labelled data, our method is more likely to offer a consistent performance boost than other pre-training approaches.

Our results give no definitive answer regarding which algorithm works best for a specific task or dataset.
Since there is no way (at present) to accurately measure the quality of features produced from a pre-training algorithm, the only way to quantitatively evaluate features is to compare downstream performance.
Downstream performance is highly variable and without knowing what makes a `good feature' for these tasks, we are forced to proceed with development following a mostly trial-and-error methodology.
Therefore, during algorithm design for a new dataset, experimentation may be required to determine which method can improve downstream performance the most.
However, our method may be a good first candidate, given its relatively large performance gain, consistency across dataset sizes, and the simplicity gained by requiring individual scans only.

\subsection{Training Speed-up}
\label{sec:results:speedup}

\begin{figure}[t!]
  \centering
  \includegraphics[width=8.25cm]{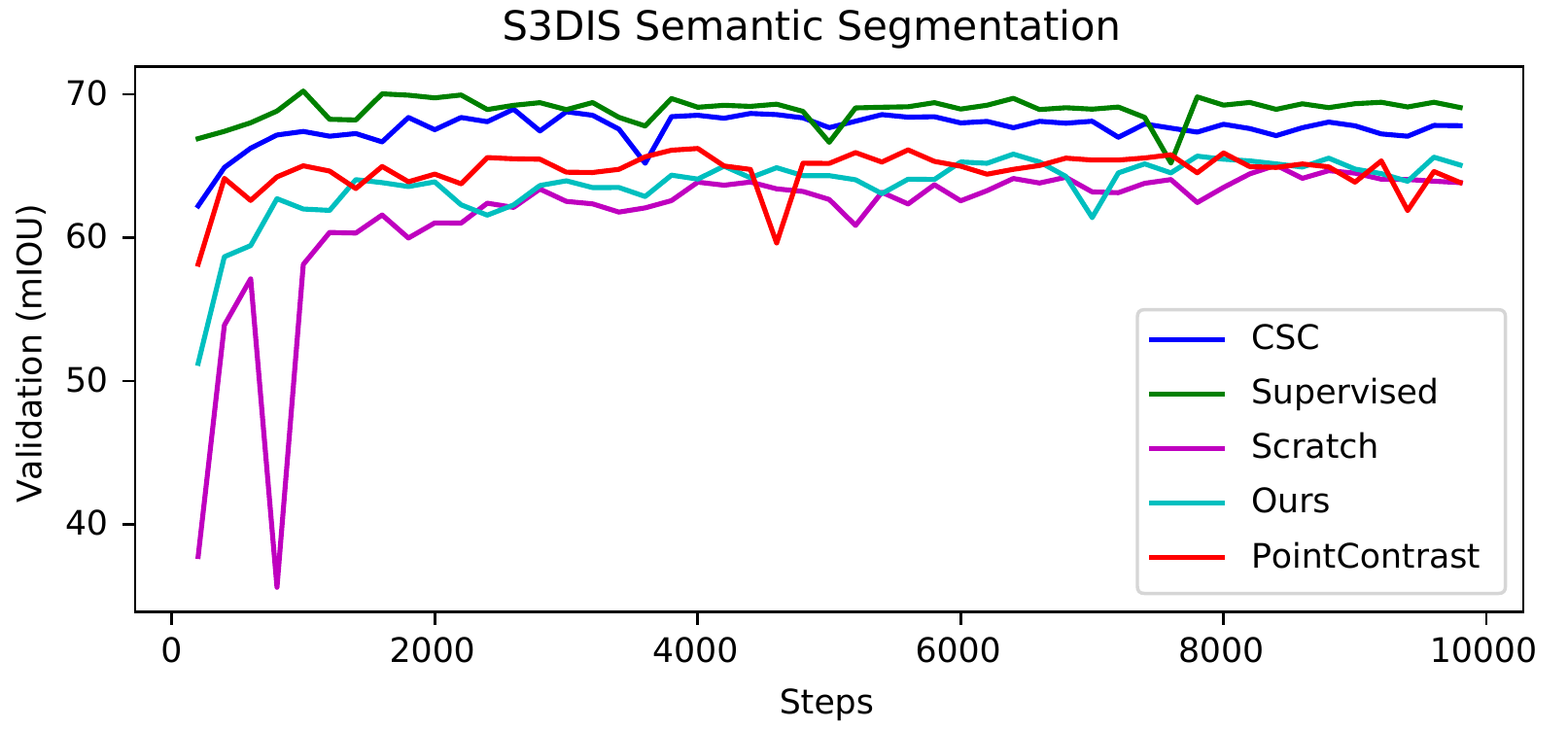}
  \caption{Comparison of validation performance over training steps for models pre-trained on ScanNet and fine-tuned for semantic segmentation on S3DIS.}
  \label{fig:validationComparison}
  \vspace{-6mm}
\end{figure}

\Cref{fig:validationComparison} shows the validation curves versus number of training steps and compares training from scratch to other training methods.
All techniques show an immediate improvement early on in the training cycle.
This result is potentially valuable during development for downstream tasks: the final performance can be determined roughly within the first 2,000 cycles, instead of 20,000.
For S3DIS, for example, algorithms like PointContrast can reach a performance level that is within 1 mIOU of the final from-scratch performance within just 1,000 cycles.
The early improvement observation should drastically reduce training and development time.

\section{Conclusion}
\label{sec:conclusion}

In this paper, we presented a method for transferring self-supervised features derived from dense images to models that operate on sparse point clouds.
We showed how pre-training affects performance on three different downstream tasks and three different datasets.
Our experiments yielded several key findings.
First, we found that the performance of existing methods was inconsistent relative to our own;
specifically, our approach had a smaller variance in accuracy when training on different amounts of labelled data for a given dataset.
Second, we found, for all methods we evaluated, that using the same data for pre-training and training yields no performance improvement.
Instead, larger amounts of data (or a different dataset entirely) are required for pre-training to improve downstream performance.
Finally, we showed that incorporating visual data into the pre-training procedure is a viable strategy to reduce or eliminate the need for registered point clouds during pre-training, improving the scalability of our method (compared with others that require multiple, registered scans).
Given this improved scalability and more consistent performance improvement across a number of downstream tasks, we believe that our method is a reasonable starting point when deciding on a pre-training strategy for 3D data.

\bibliographystyle{IEEEtran}
\bibliography{venues_abbr,2023-janda-self-supervised-crv_short}

\end{document}